\documentclass[10pt,twocolumn,letterpaper]{article}

\usepackage{iccv}
\usepackage{times}
\usepackage{epsfig}
\usepackage{graphicx}
\usepackage{amsmath}
\usepackage{amssymb}
\usepackage{subcaption}
\usepackage{float}
\usepackage{booktabs}
\usepackage{siunitx}
\usepackage{enumitem}
\usepackage{caption}
\usepackage{xcolor}

\usepackage[pagebackref=true,breaklinks=true,letterpaper=true,colorlinks,bookmarks=false]{hyperref}
\usepackage[capitalize]{cleveref}
\crefname{section}{Sec.}{Secs.}
\Crefname{section}{Section}{Sections}
\Crefname{table}{Table}{Tables}
\crefname{table}{Tab.}{Tabs.}

\iccvfinalcopy 

\ificcvfinal\fi

\begin{document}

\title{\textbf{TexLiDAR: Automated Text Understanding for Panoramic LiDAR Data}}

\author{
\begin{tabular}[t]{ccc}
Naor Cohen & Roy Orfaig & Ben-Zion Bobrovsky \\
{\tt\footnotesize naorcohen1@mail.tau.ac.il} & {\tt\footnotesize royorfaig@tauex.tau.ac.il} & {\tt\footnotesize bobrov@tauex.tau.ac.il}
\end{tabular}
\\
{\small School of Electrical Engineering, Tel-Aviv University} \\
}
\maketitle
\maketitle

\begin{abstract}
    Efforts to connect LiDAR data with text, such as LidarCLIP, have primarily focused on embedding 3D point clouds into CLIP’s text-image space. However, these approaches rely on 3D point clouds, which present challenges in encoding efficiency and neural network processing. With the advent of advanced LiDAR sensors like Ouster's OS1, which, in addition to 3D point clouds, produce fixed-resolution depth, signal, and ambient panoramic 2D images, new opportunities emerge for LiDAR-based tasks.
    
    In this work, we propose an alternative approach to connect LiDAR data with text by leveraging 2D imagery generated by the OS1 sensor instead of 3D point clouds. Using the Florence 2 large model in a zero-shot setting, we perform image captioning and object detection. Our experiments demonstrate that Florence 2 generates more informative captions and achieves superior performance in object detection tasks compared to existing methods like CLIP. By combining advanced LiDAR sensor data with a large pre-trained model, our approach provides a robust and accurate solution for challenging detection scenarios, including real-time applications requiring high accuracy and robustness.

For more details, visit our GitHub repository:
    \newline
    \url{https://github.com/AIROTAU/TexLiDAR}
    
\end{abstract}

\section{Introduction}

The integration of LiDAR technology with deep learning has advanced significantly, driven by innovations in both sensor hardware and machine learning models. Traditional LiDAR systems produce 3D point clouds that offer rich spatial information but are computationally intensive to process. Recent work, such as LidarCLIP~\cite{hess2024lidarclip}, has attempted to bridge the gap between LiDAR data and textual descriptions by embedding 3D point clouds into the CLIP framework. However, this approach has limitations: it requires aligning point clouds with a camera’s field of view, discarding the broader 360-degree spatial context, and relies on CLIP \cite{CLIP}, which provides abstract and less detailed outputs for tasks like image captioning.

The Ouster OS1 LiDAR sensor addresses some of these challenges by outputting high-resolution depth, signal, and ambient images that are spatially coherent and cover the full 360-degree field of view~\cite{OusterOS1}. These structured 2D images retain the spatial richness of LiDAR data while enabling direct processing with advanced deep learning models. Compared to 3D point clouds, they offer better compatibility with existing neural network architectures, allowing for efficient and scalable solutions.

\begin{figure}[H]
    \centering
    \begin{subfigure}[t]{0.45\textwidth}
        \centering
        \includegraphics[width=\linewidth]{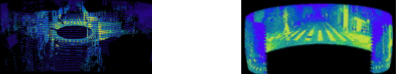}
        \caption{3D Point cloud}
        \label{fig:bottom_view}
    \end{subfigure}
    \hfill
    \begin{subfigure}[t]{0.45\textwidth}
        \centering
        \includegraphics[width=\linewidth]{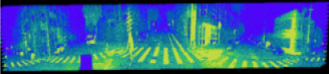}
        \caption{Panoramic Image}
        \label{fig:top_view}
    \end{subfigure}

    \caption{The Ouster OS1 sensor offers high-resolution depth, signal, and ambient images with a 360-degree field of view, ideal for lidar-based tasks. Its perfect 1:1 spatial correspondence ensures each 2D pixel maps directly to a 3D point without resampling, reducing noise, artifacts, and computational load while enhancing the accuracy of 2D and 3D perception integrations~\cite{OusterOS1}.}
    \label{fig:os1_visualization}
\end{figure}

\begin{figure*}[htbp]
    \centering
    \includegraphics[scale=0.3]{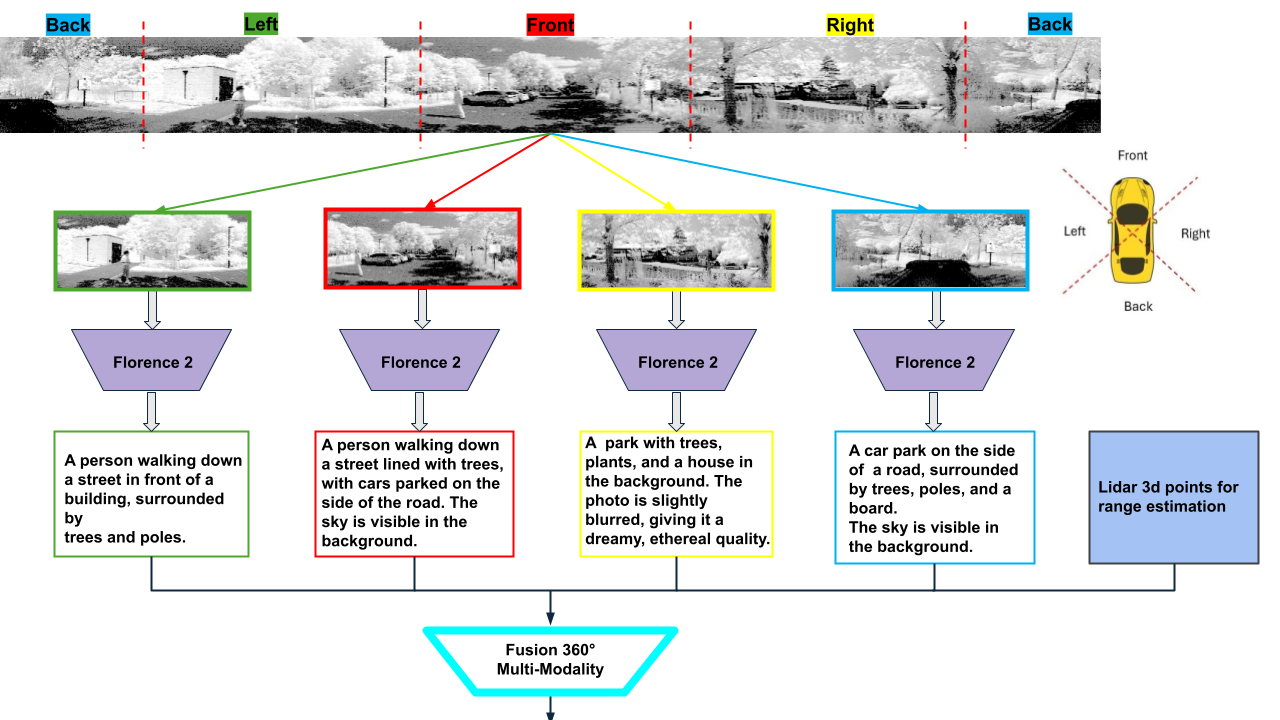} 
    \caption*{
        \texttt{\textcolor{blue}{
            Looking towards the left, the image shows a \textcolor{red}{person}\textcolor{black}{(4.9[m], -101.7\textdegree)} walking down a \textcolor{green}{street} in front of a \textcolor{purple}{building}, surrounded by \textcolor{brown}{trees} and \textcolor{orange}{poles}. 
            From the front perspective, the image depicts a \textcolor{red}{person }\textcolor{black}{(6.6[m],-29.8\textdegree)}
             walking down a \textcolor{green}{street} lined with \textcolor{brown}{trees}, with \textcolor{gray}{cars}{\textcolor{black}{ (11[m],-17.75\textdegree)}} parked on the side of the \textcolor{green}{road}. 
            The \textcolor{cyan}{sky} is visible in the background. 
            As seen from the right, the image features a \textcolor{green}{park} with \textcolor{brown}{trees}, \textcolor{brown}{plants}, and a \textcolor{purple}{house} in the background. 
            The photo is slightly blurred, giving it a dreamy, ethereal quality. 
            From the back viewpoint, we see a \textcolor{gray}{car }{\textcolor{black}{(0.8[m],-176.5\textdegree)}}
             parked on the side of a \textcolor{green}{road}, surrounded by \textcolor{brown}{trees}, \textcolor{orange}{poles}, and a \textcolor{blue}{board}. 
            The \textcolor{cyan}{sky} is visible in the background. 
        }}
    }
 
    \caption{
    Workflow for Processing 360-Degree LiDAR Images. 
    The phanoramic LiDAR image is divided into four 90-degree segments: right, left, front, and back. Each segment is processed independently by Florence 2 for image captioning. The outputs are then merged to generate a comprehensive understanding of the entire scene. The final scene description also provides positions for relevant objects (range and angle) relative to the camera (in black).
    }

    \label{fig:workflow}
\end{figure*}

In this paper, we propose leveraging the Florence 2 large model~\cite{Xiao_2024_CVPR} in a zero-shot setting to perform \textbf{image captioning} and \textbf{object detection} directly on the 2D images generated by the OS1 sensor. Florence 2’s ability to handle diverse visual tasks and generate detailed, contextually rich outputs makes it an ideal choice for lidar-based applications. By bypassing the complexities of 3D point cloud processing and fully utilizing the 360-degree data, our approach achieves more informative and accurate results compared to traditional methods.

This paper makes the following contributions:
\begin{itemize}
    \item {We introduce a framework for leveraging advanced lidar sensor data with the Florence 2 model, bypassing the need for 3D point cloud processing.}
    \item {We demonstrate the effectiveness of our approach in performing image captioning and object detection tasks, achieving more detailed and accurate results compared to existing methods.}
    \item {We highlight the ability of our method to utilize the full 360-degree field of view provided by Ouster sensors, overcoming the limitations of alignment-based approaches like LidarCLIP~\cite{hess2024lidarclip}.}
    \item {In addition to captioning and detection, we also demonstrate the capability of estimating the angle and distance of objects relative to the sensor, using the paired point cloud data for distance calculations.}
\end{itemize}

\section{Related Work}

Lidar technology plays an essential role in autonomous systems by providing high-resolution depth and spatial information. Traditional lidar data processing primarily focuses on 3D point clouds, which deliver detailed environmental data but are computationally expensive and require complex preprocessing. More recent approaches have integrated lidar data with deep learning models, such as LidarCLIP \cite{hess2024lidarclip}, which embed 3D point clouds into a shared space alongside images and text. However, these methods are constrained by the need to align point clouds with the camera's field of view, limiting their ability to fully leverage lidar’s 360-degree spatial context.

The Ouster OS1 lidar sensor \cite{OusterOS1} offers high-resolution 2D images that are spatially correlated and cover the entire 360-degree field of view. These structured images align more naturally with deep learning models, overcoming the challenges associated with processing 3D point clouds. By directly processing 2D lidar images, new possibilities arise for lidar-based tasks, such as image captioning and object detection, without the need for alignment or transformation to the camera's perspective.

Recent advances in large-scale multimodal models, such as Florence 2 \cite{Xiao_2024_CVPR}, have significantly expanded the potential applications of lidar data. Florence 2’s zero-shot capabilities enable it to perform a wide range of visual tasks—image captioning, object detection, and scene understanding—without additional training. This makes Florence 2 particularly well-suited for processing 2D lidar images, yielding more detailed and contextually relevant results compared to earlier models like CLIP \cite{CLIP}.

In contrast to LidarCLIP, which operates primarily on 3D point clouds and requires alignment with images, our approach utilizes the full 360-degree field of view provided by the Ouster OS1 sensor. By leveraging Florence 2 for zero-shot image captioning and object detection on 2D lidar images, we eliminate the need for alignment and preprocessing, yielding more accurate and insightful results for lidar-based tasks. Unlike LidarCLIP, which decodes lidar points into vectors similar to its handling of paired images \cite{hess2024lidarclip}, our approach facilitates richer, more nuanced interpretations of the data. Furthermore, future models like CLIPCap \cite{mokady2021clipcapclipprefiximage}, which combine CLIP with GPT-2 \cite{GPT2} for image captioning, often produce less informative captions. This is due to CLIP's more limited image captioning capabilities, in contrast to the more robust and contextually aware performance of Florence 2 on lidar data.

\begin{figure*}[htbp]
    \captionsetup{justification=centering, position=above}  
    \caption{\textbf{An example of our image captioning output and object detection results using 90${^\circ}$ imagery.}}
    \centering      
    \begin{minipage}[b]{0.95\linewidth}  
        \centering
        \includegraphics[width=\linewidth]{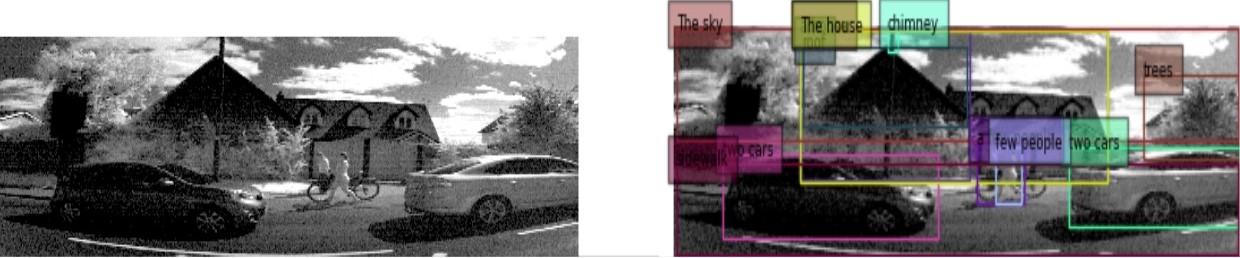}
    \caption*{
        \texttt{\textcolor{blue}{
            \textcolor{gray}{Two cars} parked on the side of a \textcolor{green}{road}, with a \textcolor{red}{person} riding a \textcolor{orange}{bicycle} in the foreground. 
            In the background, there are \textcolor{purple}{houses}, \textcolor{brown}{trees}, and a \textcolor{cyan}{sky} with \textcolor{teal}{clouds}.
        }}
    }
    \end{minipage}
    \begin{minipage}[b]{0.95\linewidth}  
        \centering
        \includegraphics[width=\linewidth]{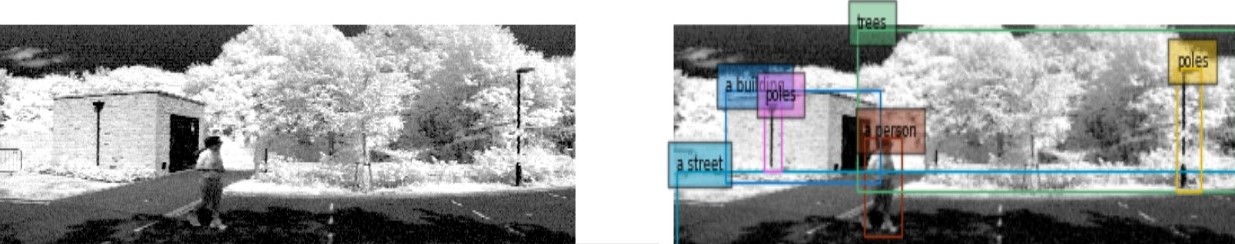}
    \caption*{
        \texttt{\textcolor{blue}{
            A \textcolor{red}{person} walking down a \textcolor{green}{street} in front of a \textcolor{purple}{building}, surrounded by \textcolor{brown}{trees} and \textcolor{orange}{poles}.
        }}
    }
    \end{minipage}
    
    \label{fig:ouster_results}
\end{figure*}

\section{Methodology}

In this work, we focus on the Drular dataset \cite{li21durlar}, specifically on ambient images, which provide detailed grayscale representations of the environment. These ambient images, with a resolution of 2048x128 pixels, capture the full 360-degree field of view from the Ouster OS1 lidar sensor \cite{OusterOS1}. These images contain complex environmental features but are not free from noise, as they stem directly from lidar sensor measurements.

The Ouster OS1 lidar sensor \cite{OusterOS1} generates structured 2D images depth, signal, and ambient that are spatially coherent and cover the full 360-degree field of view. These images are computationally efficient and highly compatible with modern deep learning models, such as Florence 2 \cite{Xiao_2024_CVPR}, which are designed for 2D image data. Unlike traditional 3D point clouds, which require complex preprocessing (e.g., transforming point clouds into a camera coordinate system or filtering out irrelevant points), 2D lidar images can be processed directly by deep learning models. This eliminates the need for extensive data transformation and reduces computational overhead, making them ideal for real-time applications with improved scalability.

In order to leverage the full potential of the 360-degree lidar data, it is crucial to feed the images into Florence 2 \cite{Xiao_2024_CVPR} in a way that preserves spatial coherence. Simply inputting the full 2048x128 image would cause Florence 2 to treat it as one large image, potentially leading to a loss of spatial understanding of the entire 360-degree scene. To address this, we divide the 360-degree image into four segments: right, left, front, and back. Each segment covers a 90-degree section of the original image, preserving the spatial layout of the scene.

Florence 2's zero-shot capabilities \cite{Xiao_2024_CVPR} make it particularly well-suited for processing structured 2D image inputs, such as the segmented images derived from 360-degree lidar data. Each segmented image, representing a 90-degree field of view, is processed independently by Florence 2, which identifies key features, objects, and their relationships within each segment. Leveraging its design for diverse visual tasks without additional training, Florence 2 performs image captioning and object detection on these inputs. The predictions from all four segments are then merged to form a comprehensive understanding of the entire 360-degree scene.

In addition to image captioning and object detection, we leverage point cloud data paired with each ambient image to estimate the angle and distance of detected objects relative to the sensor. The point cloud provides precise distance measurements, while the 360-degree nature of lidar data enables accurate angular localization. This enhances scene understanding and provides richer, more informative outputs for real-world applications.

To achieve this, we use the Florence v2 model to detect objects in ambient images and generate bounding boxes (BBs). Each bounding box provides the pixel coordinates of the detected object, and we use its center as the reference point for our calculations.

Using the distance image derived from LiDAR data, where each pixel represents the computed distance based on the horizontal and vertical angles of the LiDAR points, we calculate the distance for each LiDAR point as:

\[
\text{distance} = \sqrt{x^2 + y^2}
\]

where \(x\) and \(y\) are the real-world coordinates of the LiDAR point. The distance image is then constructed by mapping each LiDAR point to its corresponding pixel in the image, where each pixel holds the computed distance value.

Given the center of the bounding box at pixel coordinates \( (u_{\text{BB}}, v_{\text{BB}}) \), the object's distance from the sensor is directly obtained by extracting the corresponding pixel value from the distance image:

\[
\text{distance} = \text{distance}_{\text{image}}(u_{\text{BB}}, v_{\text{BB}})
\]

where \( \text{distance}_{\text{image}}(u_{\text{BB}}, v_{\text{BB}}) \) represents the value of the distance at the pixel corresponding to the bounding box center in the distance image.

To estimate the angular position of the object relative to the sensor, we compute:

\[
\text{angle} = 360 \times \frac{u_{\text{BB}} - W/2}{W}
\]

where \( u_{\text{BB}} \) is the horizontal coordinate of the bounding box center in the image, and \( W \) is the image width.

An example of a distance image, extracted from the 3D points in the point cloud, is shown below. Given the coordinates of the bounding box center in the ambient image, we can extract the corresponding matching coordinates in the distance image. Using camera-LiDAR calibration, we map the 2D image coordinates to the 3D point cloud and estimate the distance from the sensor.

The image has a width of \( W \) pixels and covers a 360-degree field of view. The center of the image corresponds to \( 0^\circ \), with the leftmost edge representing \( -180^\circ \) and the rightmost edge representing \( +180^\circ \). Using the formula above, we can map the horizontal pixel coordinate \( u_{\text{BB}} \) to its corresponding angle.

By combining object detection with point cloud data, we obtain accurate spatial localization of objects, significantly enhancing contextual awareness and scene interpretation.

This approach highlights the strength of Florence 2 in directly processing 2D lidar images to generate rich, context-aware outputs. Unlike methods such as LidarCLIP \cite{hess2024lidarclip}, which require alignment or complex preprocessing of lidar data, Florence 2 bypasses these steps, providing more accurate and informative results. This makes it an effective tool for lidar-based tasks, including image captioning and object detection.

The following key points summarize the advantages of combining lidar sensor data with large multimodal models like Florence 2:

\section*{Key Points} 
\begin{itemize} 
    \item \textbf{Zero-Shot Capability:} Florence 2 excels in zero-shot tasks, enabling effective interpretation of lidar data without specific training. 
    \item \textbf{Spatial Coherence:} Florence 2 processes spatially coherent 2D lidar images more effectively than unstructured 3D point clouds. 
    \item \textbf{Efficient Processing:} Dividing the 360-degree lidar image into smaller sections allows Florence 2 to retain global context while processing each segment independently.
    \item \textbf{Object Angle and Distance Estimation:} Leveraging the paired point cloud data with ambient images, and using object detection from Florence 2, we estimate the angle and distance of objects relative to the lidar sensor, adding another layer of contextual information.
\end{itemize}

\section{Results}
The proposed approach, leveraging the Florence 2 model for image captioning and object detection on 2D lidar images, was evaluated using data from the Ouster OS1 sensor. The sensor’s high-resolution depth, signal, and ambient images were processed to generate captions and detect objects within the scene. Figure \ref{fig:ouster_results} illustrates these results, showcasing both tasks on 90° imagery. The left side presents the captioning output, where the model describes key elements such as vehicles, pedestrians, and background structures. On the right, object detection results demonstrate accurate identification and localization.

Further analysis, shown in Figure \ref{fig:workflow}, incorporates object distance and angle estimation, enhancing spatial awareness. By segmenting the 360-degree lidar image into four 90-degree views, the model processes each independently before merging the outputs into a unified scene description. This capability is particularly valuable for applications such as autonomous navigation, robotic perception, and smart city monitoring, where precise environmental understanding is crucial.

By integrating captioning and detection within a single framework, Florence 2 enables comprehensive scene interpretation. Its robust performance across diverse environments demonstrates its potential for real-world deployment, advancing perception capabilities beyond traditional 2D image analysis.

\begin{figure*}[htbp]
    \centering
    \begin{minipage}[t]{0.45\linewidth}
        \centering
        \textbf{Reflectivity}
    \end{minipage}
    \hspace{0.2cm}
    \begin{minipage}[t]{0.45\linewidth}
        \centering
        \textbf{Ambient}
    \end{minipage}
    \vspace{0.2cm}
    \begin{minipage}[t]{0.45\linewidth}
        \centering
        \includegraphics[width=\linewidth]{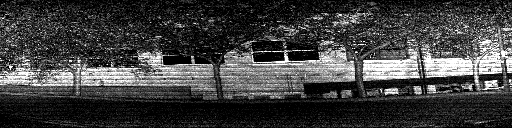}
        \caption*{\texttt{\textcolor{blue}{A \textcolor{purple}{house} at night, surrounded by \textcolor{brown}{trees} and a \textcolor{gray}{fence}. The \textcolor{purple}{house} is illuminated by the \textcolor{cyan}{moonlight}, casting a soft glow on the surrounding area, despite being partially obscured by the \textcolor{brown}{trees} and \textcolor{green}{bushes}.}}}
    \end{minipage}
    \hspace{0.2cm}
    \begin{minipage}[t]{0.45\linewidth}
        \centering
        \includegraphics[width=\linewidth]{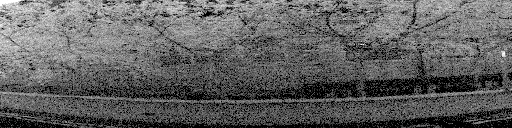}
        \caption*{\texttt{\textcolor{blue}{ A \textcolor{gray}{large rock} in the middle of a \textcolor{green}{field}, with \textcolor{brown}{trees} in the background.}}}
    \end{minipage}
    \vspace{0.5cm}
    
    \begin{minipage}[t]{0.45\linewidth}
        \centering
        \includegraphics[width=\linewidth]{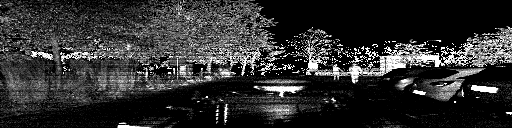}
        \caption*{\texttt{\textcolor{blue}{A \textcolor{gray}{car} parked in a \textcolor{green}{parking lot} at night, surrounded by \textcolor{brown}{trees} and \textcolor{purple}{buildings} in the background.}}}
    \end{minipage}
    \hspace{0.2cm}
    \begin{minipage}[t]{0.45\linewidth}
        \centering
        \includegraphics[width=\linewidth]{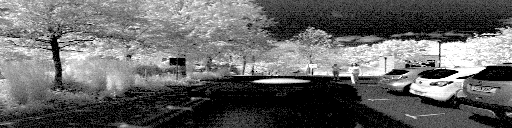}
        \caption*{\texttt{\textcolor{blue}{A \textcolor{green}{street} with \textcolor{gray}{cars} parked on the side of it, surrounded by \textcolor{brown}{trees}, \textcolor{orange}{poles}, and \textcolor{red}{people}. The \textcolor{cyan}{sky} is visible in the background.}}}
    \end{minipage}
    \caption{Examples of additional information in reflectivity vs. ambient images from the Drular dataset.}
    \label{fig:reflectivity_ambient_examples}
\end{figure*}

\section{Discussion}

The results obtained using ambient images highlight the potential of lidar-generated 2D images for tasks such as image captioning and object detection. However, the Drular dataset \cite{li21durlar} includes additional image modalities, such as reflectivity images, which capture the strength of the reflected lidar signal. These reflectivity images can provide unique insights into material properties and surface characteristics, offering a complementary perspective to ambient images.

As shown in Figure \ref{fig:reflectivity_ambient_examples}, the reflectivity images tend to reveal hidden or obscured objects that are not easily visible in the ambient versions. In the first pair, the reflectivity image highlights a house partially obscured by trees, which is barely visible in the ambient version. In the second pair, while the ambient image captures broader context, including people in the scene, the reflectivity image does not provide any additional advantages. In general, due to the black-and-white nature of the reflectivity images, the model often interprets them as night-time scenes, which can affect the model’s interpretation of the data. These examples demonstrate how reflectivity and ambient images complement each other by revealing different elements of the scene.

Exploring reflectivity images using the same Florence 2-based methodology \cite{Xiao_2024_CVPR} could further enhance performance in lidar-based tasks. For instance, the reflectivity data could improve object detection in scenarios involving low-visibility conditions or materials with distinct reflectance properties, such as metallic or glass surfaces.

Moreover, combining all available lidar modalities, including ambient, reflectivity, intensity, and range images, could lead to a powerful multimodal system. Each modality contributes unique information:
\begin{itemize}
    \item \textbf{Ambient images}: grayscale ambient imagery at 850 nm offers a high signal-to-noise ratio, improving usability in low-light conditions like dawn, dusk, or cloudy days.
    \item \textbf{Reflectivity images}: highlights material properties and surface textures based on the detected surface reflectivity.
    \item \textbf{Intensity images}: represents lidar intensity data, highlighting anomalies and strong signal returns based on detected photon counts.
    \item \textbf{Range images}: point distance is calculated using the laser pulse's time of flight, providing accurate 3D measurements and enabling precise object localization in a scene.
\end{itemize}

A multimodal approach could leverage these complementary strengths to deliver more robust and context-aware results. Florence 2’s versatility in processing diverse image types \cite{Xiao_2024_CVPR} makes it an excellent candidate for implementing such a system.

Future research could focus on developing methods to efficiently fuse these modalities, exploring the potential of multimodal systems in applications such as autonomous driving, environmental monitoring, and robotics.

\section{Conclusion}

In this paper, we have introduced a novel methodology for leveraging advanced lidar sensor data, particularly the 2D images generated by the Ouster OS1 sensor \cite{OusterOS1}, in conjunction with the Florence 2 large model \cite{Xiao_2024_CVPR} for tasks such as image captioning and object detection. By focusing on the structured, high-resolution 2D images, we bypass the complexities associated with 3D point cloud processing and demonstrate the potential of zero-shot inference for lidar-based tasks.

Our approach effectively utilizes the full 360-degree spatial context provided by the Ouster OS1 sensor, dividing the panoramic ambient images into manageable segments for processing with Florence 2. The results highlight the advantages of this method, showing detailed and contextually rich outputs that surpass traditional methods such as LidarCLIP \cite{hess2024lidarclip}.

Furthermore, we discussed the possibility of extending this approach to reflectivity images from the Drular dataset \cite{li21durlar}, emphasizing the complementary insights they can provide. We also proposed the potential of a multimodal system that integrates multiple lidar image modalities—ambient, reflectivity, intensity, and range—to enhance overall performance in various applications.

The findings underscore the versatility of the Florence 2 model \cite{Xiao_2024_CVPR} and its capacity to process diverse lidar-based image types effectively. Future work can focus on optimizing the fusion of these modalities and exploring additional applications in autonomous systems, robotics, and environmental monitoring. Our research sets the foundation for scalable, efficient, and highly informative solutions for lidar-based perception tasks.

\bibliographystyle{ieeetr}  
\bibliography{ref}  

\begin{thebibliography}{1}

\bibitem{hess2024lidarclip}
G.~Hess, A.~Tonderski, C.~Petersson, K.~{\AA}str{\"o}m, and L.~Svensson, ``Lidarclip or: How i learned to talk to point clouds,'' in {\em Proceedings of the IEEE/CVF Winter Conference on Applications of Computer Vision}, pp.~7438--7447, 2024.

\bibitem{CLIP}
A.~Radford, J.~W. Kim, C.~Hallacy, A.~Ramesh, G.~Goh, S.~Agarwal, G.~Sastry, A.~Askell, P.~Mishkin, J.~Clark, G.~Krueger, and I.~Sutskever, ``Learning transferable visual models from natural language supervision,'' in {\em Proceedings of the 38th International Conference on Machine Learning} (M.~Meila and T.~Zhang, eds.), vol.~139 of {\em Proceedings of Machine Learning Research}, pp.~8748--8763, PMLR, 18--24 Jul 2021.

\bibitem{OusterOS1}
O.~Inc., ``The camera is in the lidar: Ouster os1 sensor,'' 2022.

\bibitem{Xiao_2024_CVPR}
B.~Xiao, H.~Wu, W.~Xu, X.~Dai, H.~Hu, Y.~Lu, M.~Zeng, C.~Liu, and L.~Yuan, ``Florence-2: Advancing a unified representation for a variety of vision tasks,'' in {\em Proceedings of the IEEE/CVF Conference on Computer Vision and Pattern Recognition (CVPR)}, pp.~4818--4829, June 2024.

\bibitem{mokady2021clipcapclipprefiximage}
R.~Mokady, A.~Hertz, and A.~H. Bermano, ``Clipcap: Clip prefix for image captioning,'' 2021.

\bibitem{GPT2}
T.~Brown, B.~Mann, N.~Ryder, M.~Subbiah, J.~D. Kaplan, P.~Dhariwal, A.~Neelakantan, P.~Shyam, G.~Sastry, A.~Askell, S.~Agarwal, A.~Herbert-Voss, G.~Krueger, T.~Henighan, R.~Child, A.~Ramesh, D.~Ziegler, J.~Wu, C.~Winter, C.~Hesse, M.~Chen, E.~Sigler, M.~Litwin, S.~Gray, B.~Chess, J.~Clark, C.~Berner, S.~McCandlish, A.~Radford, I.~Sutskever, and D.~Amodei, ``Language models are few-shot learners,'' in {\em Advances in Neural Information Processing Systems} (H.~Larochelle, M.~Ranzato, R.~Hadsell, M.~Balcan, and H.~Lin, eds.), vol.~33, pp.~1877--1901, Curran Associates, Inc., 2020.

\bibitem{li21durlar}
L.~Li, K.~Ismail, H.~Shum, and T.~Breckon, ``Durlar: A high-fidelity 128-channel lidar dataset with panoramic ambient and reflectivity imagery for multi-modal autonomous driving applications,'' in {\em Proc. Int. Conf. on 3D Vision}, IEEE, December 2021.

\end{thebibliography}
\end{document}